\documentclass[lettersize,journal]{IEEEtran}
\usepackage{xcolor}
\usepackage[]{hyperref}
\hypersetup{colorlinks,breaklinks,linkcolor=blue,urlcolor=blue,anchorcolor=blue,citecolor=blue}
\usepackage{amsmath,amsfonts}
\usepackage{amsthm}

\usepackage{upgreek}
\usepackage{easyReview}
\usepackage{setspace}

\usepackage{algorithmic}
\usepackage{algorithm}
\usepackage{array}
\usepackage{textcomp}
\usepackage{stfloats}
\usepackage{url}
\usepackage{verbatim}
\usepackage{graphicx}
\usepackage{adjustbox}
\usepackage[nocompress]{cite}
\usepackage{caption}
\usepackage{subcaption}
\usepackage{bm, amssymb, fixmath,dsfont}
\let\oldnl\nl
\newcommand{\nonl}{\renewcommand{\nl}{\let\nl\oldnl}}
\usepackage{color}
\usepackage{soul}
\usepackage{multirow}
\usepackage{booktabs}
\usepackage{bbm}
\usepackage{tikz}
\usepackage{rotating}

\newtheorem{definition}{Definition}

\usepackage{circledsteps}
\usepackage{physics}
\usepackage{tabularx, makecell, setspace}
\usepackage{circledsteps}
\usepackage{physics}
\usepackage{adjustbox}
\usepackage{svg}
\usetikzlibrary{shapes.geometric, arrows, positioning}
\usepackage{arydshln}

\title{\LARGE \bf
Policy Optimality Measurement for Multi-Vehicle Decision-Making: From Extrinsic Indicators to Intrinsic Quality
}

\author{
Ye Han, Lijun Zhang$^*$, Dejian Meng  
\thanks{Ye Han, Lijun Zhang, Dejian Meng are with the School of Automotive Studies, Tongji University, Shanghai 201804, China.
        {\tt\small \{hanye\_leohancnjs, tjedu\_zhanglijun, mengdejian\}@tongji.edu.cn}}%
\thanks{$^*$Corresponding author: Lijun Zhang}
}

\begin{document}

\maketitle
\thispagestyle{empty}
\pagestyle{empty}

\begin{abstract}

Evaluating Multi-Agent Reinforcement Learning (MARL) policies in autonomous driving fundamentally relies on extrinsic statistical indicators (e.g., reward curves and success rates), which often mask intrinsic policy degradation and algorithmic blind spots. To break this black-box evaluation, this letter proposes a novel information-theoretic diagnostic framework. By leveraging a fully converged Monte Carlo Tree Search (MCTS) as an asymptotic oracle, we establish a theoretical ground-truth baseline distribution. We formulate a bounded policy optimality score ($\mathcal{M}_{opt}$) using the forward KL divergence to rigorously penalize fatal collaborative omissions. Crucially, we semantically decouple this metric into lateral and longitudinal dimensions, creating a granular ``semantic microscope''. Extensive spatial and temporal diagnostics on state-of-the-art MARL architectures and exploration mechanisms demonstrate that our framework conclusively exposes hidden directional biases, identifies temporal average-policy traps, and transforms heuristic hyperparameter tuning into a visually trackable trajectory optimization. This framework establishes a rigorous, model-agnostic standard for benchmarking intrinsic multi-agent policy quality.

\end{abstract}

\section{Introduction}
\label{sec:introduction}

The coordination of Connected and Automated Vehicles (CAVs) in dense, mixed-traffic environments is fundamentally formulated as a partially observable Markov Game \cite{hua2024multi, wang2024multi}. Due to the continuous state spaces, high-dimensional joint action spaces, and the dynamic number of interacting vehicles, traditional rule-based and centralized optimization methods often fall short \cite{chen2021graph, xu2024multi}. Consequently, Multi-Agent Reinforcement Learning (MARL), particularly under the Centralized Training with Decentralized Execution (CTDE) paradigm, has emerged as the frontier solution for multi-vehicle collaborative decision-making \cite{hua2024multi, guillen2022multi}. 
To resolve the complex non-Euclidean interactive topologies among CAVs and human-driven vehicles (HDVs), recent studies have intensely focused on architectural innovations. Advanced spatial-temporal representation networks, such as Graph Neural Networks (GNNs) \cite{chen2021graph, zhang2023spatial} and Transformer-based attention mechanisms \cite{xu2024multi}, have been heavily investigated. Concurrently, a plethora of specialized mechanisms—ranging from heuristic action masking \cite{guo2024heuristic, chen2023deep}, control barrier functions (CBF) \cite{han2022multi, zhang2023spatial}, to multi-objective reward shaping \cite{saleem2024multi, wang2023multi}—have been proposed to enhance policy convergence and empirical safety. 

Despite the proliferation of MARL architectures, the field has hit a severe bottleneck in policy evaluation \cite{zhou2020smarts}. The vast majority of current autonomous driving MARL research fundamentally relies on \textit{extrinsic statistical indicators} to benchmark algorithmic superiority. These metrics predominantly consist of macro-level reward curves and post-simulation statistics, such as task success rates, collision rates, travel times, and traffic throughput \cite{guillen2022multi, xu2024multi, saleem2024multi, guo2025opencda}. However, these superficial metrics mask the intrinsic degradation of the learned policies, leading to three critical diagnostic blind spots.

First, the reliance on aggregated scalar rewards is intrinsically flawed for measuring multi-agent synergy, often leading to \textit{reward hacking} \cite{yan2024policy, li2025hierarchical}. Hand-crafted weighted summations of safety, efficiency, and comfort inevitably induce agents to exploit environmental artifacts rather than learning genuine collaborative intelligence \cite{liu2025rrm, abouelazm2024review}. Consequently, a converged cumulative return curve merely proves that the algorithm has overfitted to a specific heuristic, rather than achieving policy optimality \cite{jordan2020evaluating}.

Second, the widespread tune-and-report paradigm creates an illusion of algorithmic progress \cite{henderson2018deep, jordan2020evaluating}. The black-box nature of deep reinforcement learning \cite{glanois2024survey} means that averaging extrinsic metrics over multiple episodes obscures the severe fragility and variance of the underlying probability distributions. An algorithm might occasionally yield high macro-returns but completely fail to guarantee structural verifiability in edge cases.

Furthermore, high task success rates or collision-free statistics do not necessarily equate to advanced collaborative intelligence \cite{zhou2020smarts}. Empirical evaluations show that a 100\% success rate can sometimes be achieved by algorithms that degenerate into overly conservative, static average behaviors (e.g., refusing to negotiate and simply waiting), severely degrading system-wide efficiency \cite{chen2021graph}. Even under high-density benchmarks, models boasting high throughput may still harbor significant collision risks due to fatal spatial misjudgments \cite{guo2025opencda}. Extrinsic metrics fail to explain \textit{why} a policy succeeded or mathematically pinpoint its algorithmic blindness—whether the model is specifically failing in lateral lane-change planning or longitudinal velocity coordination \cite{wang2023multi, chen2023deep}.

To break this black box of MARL evaluation, a transition from extrinsic indicators to intrinsic information-theoretic metrics is urgently required. To evaluate the intrinsic quality of MARL probability distributions rather than their empirical outcomes, we need a mathematically rigorous baseline. Since both MARL and heuristic tree search fundamentally approximate the same Bellman optimality objective, Monte Carlo Tree Search (MCTS) can serve as an asymptotic oracle \cite{swiechowski2023monte}. Utilizing MCTS as an expert demonstrator to evaluate and guide neural network policies has achieved profound success in complex zero-sum and multi-agent games \cite{silver2017mastering, helfenstein2024checkmating}. By extracting the optimal value landscape from a fully converged MCTS and applying a Boltzmann mapping, we can project the absolute value advantages into a theoretical baseline probability distribution \cite{swiechowski2023monte}.

With the theoretical baseline established, we transition from scalar reward evaluations to information-theoretic distribution comparisons. Information theory, particularly the Kullback-Leibler (KL) divergence, has long provided the mathematical orthodoxy for bounding policy evolution and quantifying distribution shifts in foundational RL algorithms \cite{schulman2015trust, schulman2017proximal}. Recent advancements further validate the use of policy divergence to constrain vast exploration spaces \cite{vinyals2019grandmaster} and explicitly quantify multi-agent behavioral heterogeneity \cite{dou2024measuring}. Crucially, by adopting the forward KL divergence, our framework leverages its inherent mode-covering asymmetry \cite{chan2022greedification}. This heavily penalizes the MARL policy if it assigns near-zero probabilities to critical collaborative maneuvers highlighted by the MCTS oracle. Consequently, it functions as a rigorous mathematical detector for fatal omissions \cite{chan2022greedification} in dynamic traffic negotiations, effectively bypassing the deceptive nature of extrinsic rewards.

In this paper, we propose an information-theoretic diagnostic framework that transitions MARL evaluation in autonomous driving from extrinsic statistical indicators to intrinsic policy quality. The primary contributions are summarized as follows:

\begin{itemize}
    \item \textbf{Information-Theoretic Optimality Metric Formulation:} We establish a theoretical baseline via an MCTS oracle and reconstruct the MARL joint policy under the CTDE paradigm. By leveraging the forward KL divergence, we formulate a bounded optimality score ($\mathcal{M}_{opt}$) that rigorously quantifies the intrinsic structural discrepancy of multi-agent policies.
    \item \textbf{Spatial Diagnostics via Semantic Decoupling:} We semantically decouple the holistic joint policy into distinct lateral and longitudinal optimality metrics ($\mathcal{M}_{opt, lat}$, $\mathcal{M}_{opt, lon}$). This allows the framework to act as a semantic microscope, accurately diagnosing spatial reasoning blind spots in representation network architectures and exposing the severe directional biases inherent in SOTA unconstrained exploration algorithms.
    \item \textbf{Temporal Diagnostics and Guided Optimization:} We visualize the temporal evolution of the decoupled metrics to expose average-policy and collusion traps hidden by high task success rates. Furthermore, we transform hyperparameter tuning from high-variance reward-hacking into a visually trackable trajectory optimization process in a 2D metric space, ensuring balanced collaborative exploration.
\end{itemize}
\section{Formulation of the Policy Optimality Metric}
\label{sec:metric_formulation}

\begin{figure}[htbp]
    \centering
    \includegraphics[width=\linewidth]{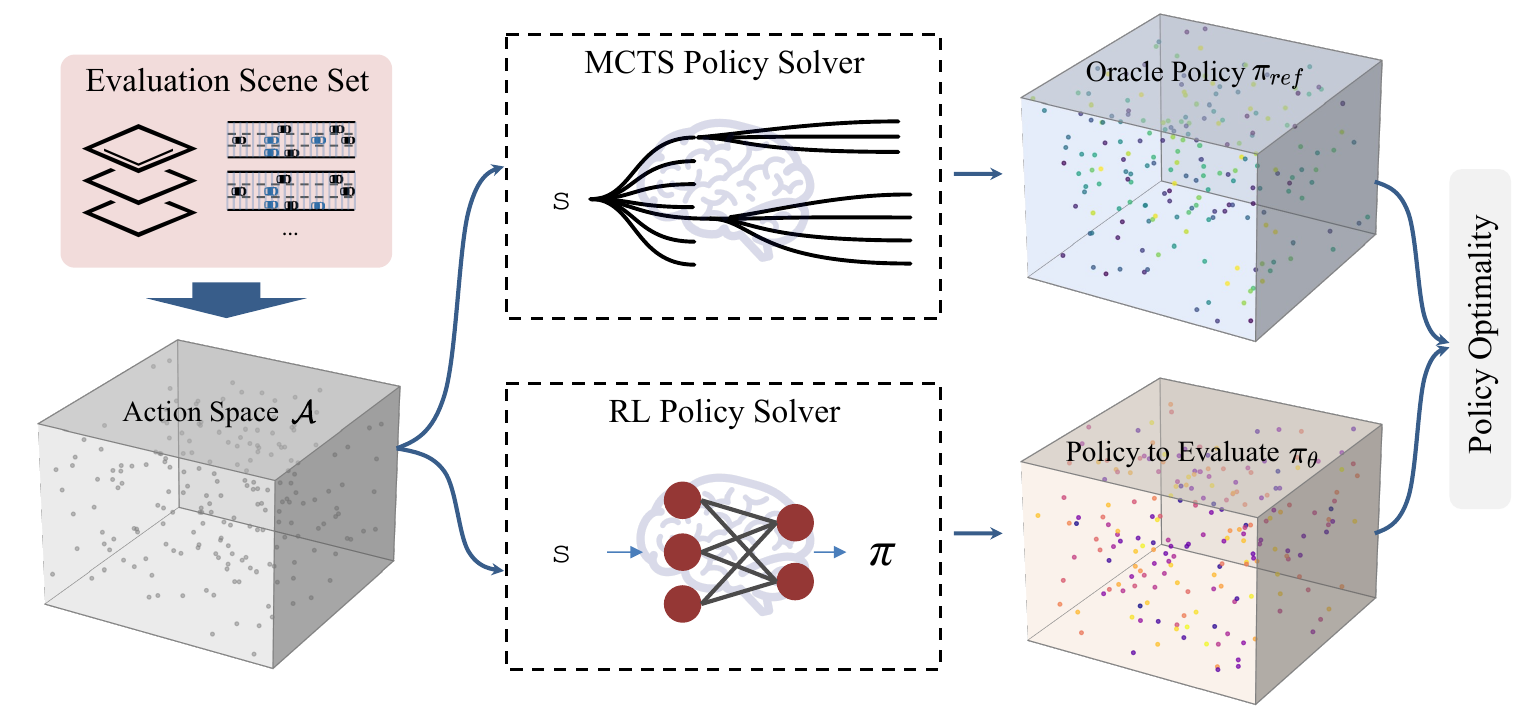}
    \caption{Overall framework for generating the policy optimality metric. The theoretical baseline $\pi_{ref}$ derived from the MCTS oracle and the evaluated policy $\pi_{\theta}$ from the MARL algorithm are projected into a unified probability measure space for structural comparison.}
    \label{fig:optimality_framework}
\end{figure}

To establish a rigorous diagnostic framework for multi-agent policies, we must first define the shared physical and mathematical space in which both heuristic search and reinforcement learning operate. The overall workflow for generating this metric is illustrated in Fig. \ref{fig:optimality_framework}.

\subsection{Multi-Vehicle Collaboration as a Markov Game}
\label{subsec:markov_game}

We formulate the multi-vehicle collaborative decision-making process in a mixed traffic environment as a fully cooperative Markov Game. This process is defined by the tuple $\langle \mathcal{N}, \mathcal{S}, \mathcal{A}, \mathcal{P}, \mathcal{R}, \gamma \rangle$, $\mathcal{N} = \{1, 2, \dots, N\}$ represents the dynamic set of controlled Connected and Automated Vehicles (CAVs) within the coordination region. $\mathcal{S}$ is the continuous state space capturing the global traffic kinematics and topology. $\mathcal{A} \triangleq \prod_{i=1}^N \mathcal{A}^{(i)}$ denotes the vast joint action space, where $\mathcal{A}^{(i)}$ is the discrete lateral and longitudinal action space of an individual CAV $i$. $\mathcal{P}: \mathcal{S} \times \mathcal{A} \times \mathcal{S} \to [0, 1]$ represents the state transition dynamics, which encapsulate the physical constraints and the highly uncertain behaviors of background Human-Driven Vehicles (HDVs). $\mathcal{R}: \mathcal{S} \times \mathcal{A} \to \mathbb{R}$ is the global reward function that evaluates the overall team performance (e.g., safety, throughput efficiency, and smoothness) resulting from the joint action $\bm{a} = (a^{(1)}, \dots, a^{(N)})$. $\gamma \in (0, 1)$ is the discount factor.

The ultimate objective of this multi-vehicle system is to discover an optimal joint policy $\bm{\pi}^*$ that maximizes the expected cumulative discounted return from any given state. This objective is strictly governed by a unique optimal joint action-value function, $Q^*(s, \bm{a})$.

\begin{definition}[Bellman Optimality Objective]
For the cooperative Markov Game, the optimal joint action-value function $Q^*(s, \bm{a})$ represents the maximum achievable expected return when taking joint action $\bm{a}$ in state $s$, and subsequently following the optimal policy. It satisfies the Bellman optimality equation:
\begin{equation}
    Q^*(s, \bm{a}) = \mathcal{R}(s, \bm{a}) + \gamma \sum_{s' \in \mathcal{S}} \mathcal{P}(s'|s, \bm{a}) \max_{\bm{a}' \in \mathcal{A}} Q^*(s', \bm{a}').
    \label{eq:bellman_opt}
\end{equation}
\end{definition}

\textbf{Remark (Theoretical Equivalence):} Equation \eqref{eq:bellman_opt} serves as the theoretical bridge between MCTS and MARL. Both approaches are essentially distinct numerical solvers striving to approximate this exact same mathematical target. MCTS attempts to derive $Q^*(s, \bm{a})$ online through asymmetric tree expansion and episodic rollouts, whereas MARL seeks to parameterize it offline using deep neural networks (e.g., via the CTDE paradigm). The existence of this shared Bellman optimality objective mathematically justifies utilizing the value estimates from a fully converged MCTS as a reliable ground-truth baseline to measure the intrinsic parameterization flaws of MARL policies.

\subsection{Establishing the Theoretical Baseline via Bellman Equivalence}
\label{subsec:baseline_establishment}

Since directly solving the Bellman optimality equation in high-dimensional continuous traffic flows is mathematically intractable, we require a robust numerical proxy.

\subsubsection{Asymptotic Optimality of MCTS}
Monte Carlo tree search, specifically when employing Upper Confidence Bounds applied to Trees (UCT), provides a theoretically guaranteed pathway to this proxy. Given a sufficient computational budget, the vast exploration of the joint action space allows the node action-value estimates $\hat{Q}_{\text{MCTS}}^*(s, \bm{a})$ to asymptotically converge to the true Bayesian optimal value $Q^*(s, \bm{a})$. By deploying a highly optimized MCTS solver (e.g., with parallel update pruning mechanisms\cite{han2025mcts}) as an oracle, we can extract the ground-truth value landscape of the cooperative game for any critical traffic state $s$.

\subsubsection{Boltzmann Distribution Mapping}
While MCTS yields absolute value estimates, MARL algorithms typically output stochastic policy distributions (probabilities). To enable a direct mathematical comparison using information-theoretic metrics, the absolute value domain of MCTS must be rigorously mapped into a probability measure space.

We achieve this by projecting the MCTS value landscape into a target probability distribution using a Boltzmann (softmax) formulation.

\begin{definition}[Theoretical Baseline Policy]
For any given traffic state $s$, the theoretical baseline policy distribution $\pi_{ref}(\cdot|s)$ over the joint action space $\mathcal{A}$ is defined as:
\begin{equation}
	\pi_{ref}( \bm{a} | s ) \triangleq \frac{\exp\left(\hat{Q}_{\text{MCTS}}^*(s, \bm{a}) / \kappa\right)}{\sum_{\bm{a}' \in \mathcal{A}} \exp\left(\hat{Q}_{\text{MCTS}}^*(s, \bm{a}') / \kappa\right)},
	\label{eq:boltzmann_baseline}
\end{equation}
where $\kappa > 0$ is the temperature parameter.
\end{definition}

\textbf{Remark (Engineering Semantics of Temperature):} The mapping in Equation \eqref{eq:boltzmann_baseline} transforms the absolute advantage relationships of the $Q$-values into relative probabilities. The temperature parameter $\kappa$ controls the smoothness of this projection. Instead of applying a hard \textit{argmax} that collapses the baseline into a one-hot distribution, an appropriate $\kappa$ preserves the probabilities of high-value, secondary collaborative actions. This is crucial in dynamic multi-vehicle environments, where multiple near-optimal joint maneuvers (e.g., yielding vs. accelerating) might safely resolve a traffic conflict, and punishing a MARL agent for choosing a valid secondary strategy would be unreasonable.

\subsection{Joint Policy Reconstruction for the CTDE Paradigm}
\label{subsec:joint_policy_reconstruction}

To rigorously evaluate a multi-agent reinforcement learning algorithm, its decision output must be aligned with the probability measure space of the theoretical baseline. However, prevailing MARL algorithms typically adopt the Centralized Training with Decentralized Execution (CTDE) paradigm to bypass the curse of dimensionality. Depending on the specific MARL architecture, we provide two mathematically consistent pathways to reconstruct the joint policy $\pi_{\theta}(\bm{a}|s)$.

\subsubsection{General Formulation: Independent Marginalization}
In standard CTDE frameworks (e.g., independent learners or actor-critic methods like MAPPO), agents do not explicitly output a joint policy. Instead, each CAV $i$ maintains a local utility network producing a decentralized marginal probability distribution $\pi_{\theta}^{(i)}(a^{(i)}|\tau^{(i)})$ based solely on its local history $\tau^{(i)}$.

\textbf{Assumption 1 (Conditional Independence of Execution):} During the decentralized execution phase, the action selection of each individual CAV is conditionally independent of others, given its own local observation.

\begin{definition}[General Reconstructed Joint Policy]
Under Assumption 1, the evaluated joint policy distribution for a global traffic state $s$ is reconstructed as the product of the individual marginal policies:
\begin{equation}
	\pi_{\theta}(\bm{a}|s) \triangleq \prod_{i=1}^{N} \pi_{\theta}^{(i)}(a^{(i)}|\tau^{(i)}),
	\label{eq:joint_policy_general}
\end{equation}
where $\bm{a} = (a^{(1)}, \dots, a^{(N)})$ represents the sampled joint maneuver.
\end{definition}

\begin{figure}[htbp]
    \centering
    \includegraphics[width=\linewidth]{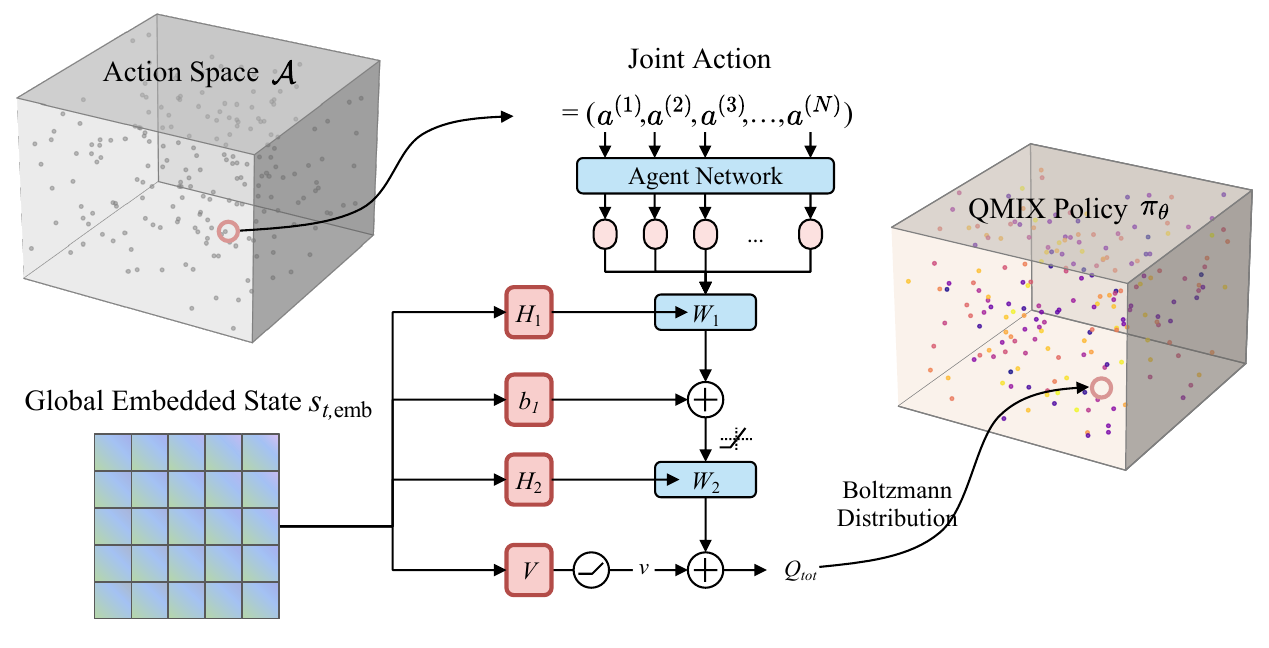}
    \caption{Synergistic joint policy reconstruction for value decomposition networks. By iterating through the joint action space $\mathcal{A}$ and leveraging the centralized mixing network, the global action-value landscape $Q_{tot}$ is extracted to formulate a high-fidelity joint probability distribution.}
    \label{fig:qmix_reconstruction}
\end{figure}

\subsubsection{Advanced Formulation for Value Decomposition}
While Equation \eqref{eq:joint_policy_general} guarantees universality, it may inadvertently discard the complex, non-linear collaborative values learned during centralized training for algorithms explicitly utilizing a mixing network (e.g., QMIX, VDN). As illustrated in Fig. \ref{fig:qmix_reconstruction}, for such value decomposition architectures, we can bypass the independence assumption to achieve a higher-fidelity reconstruction.

By iterating through the entire joint action space $\mathcal{A}$, we extract the corresponding local utilities and feed them, alongside the global state $s$, into the trained mixing network to reconstruct the complete global action-value landscape $Q_{tot}(s, \bm{a})$. 

\begin{definition}[Synergistic Reconstructed Joint Policy]
For value decomposition networks, the joint policy distribution is derived directly by applying the Boltzmann projection to the holistic mixing network output:
\begin{equation}
	\pi_{\theta}(\bm{a}|s) \triangleq \frac{\exp\left(Q_{tot}(s, \bm{a}) / \kappa\right)}{\sum_{\bm{a}' \in \mathcal{A}} \exp\left(Q_{tot}(s, \bm{a}') / \kappa\right)}.
	\label{eq:joint_policy_qtot}
\end{equation}
\end{definition}

\textbf{Remark (Universality and Fidelity):} By providing both the general marginalization pathway and the advanced synergistic pathway, our diagnostic framework guarantees universal applicability across any MARL architecture. It effectively lifts isolated, decentralized decisions back into the global joint probability space, ensuring strict commensurability with the MCTS baseline $\pi_{ref}(\bm{a}|s)$ over the domain $\mathcal{A}$.

\subsection{Information-Theoretic Optimality Formulation}
\label{subsec:information_theoretic_formulation}

With the theoretical baseline $\pi_{ref}(\cdot|s)$ and the reconstructed MARL policy $\pi_{\theta}(\cdot|s)$ strictly aligned within the same joint probability space, we can now mathematically quantify their structural discrepancy. Information theory, specifically the Kullback-Leibler (KL) divergence, provides a rigorous tool to measure the information loss incurred when using a parameterized policy to approximate the true optimal distribution.

\subsubsection{Policy Divergence via Kullback-Leibler}

\begin{definition}[Policy Optimality Distance]
For a given traffic state $s$, the policy optimality distance $\mathcal{D}_{opt}$ of the evaluated MARL policy $\pi_{\theta}$ relative to the theoretical baseline $\pi_{ref}$ is defined as the forward KL divergence from $\pi_{\theta}$ to $\pi_{ref}$:
\begin{equation}
	\begin{aligned}
		\mathcal{D}_{\text{opt}}(\pi_{\theta}|s) &\triangleq D_{\text{KL}}\left(\pi_{ref}(\cdot|s) \parallel \pi_{\theta}(\cdot|s)\right) \\
		&= \sum_{\bm{a} \in \mathcal{A}} \pi_{ref}(\bm{a}|s) \log \frac{\pi_{ref}(\bm{a}|s)}{\pi_{\theta}(\bm{a}|s)}.
	\end{aligned}
	\label{eq:policy_distance_kl}
\end{equation}
\end{definition}

\textbf{Remark (Engineering Semantics of Asymmetry):} The choice of the forward KL divergence, $D_{\text{KL}}(\pi_{ref} || \pi_{\theta})$, rather than its reverse, is highly deliberate and carries profound engineering significance. Due to its inherent asymmetry, Equation \eqref{eq:policy_distance_kl} heavily penalizes the MARL algorithm if it assigns a near-zero probability to a joint maneuver that the MCTS baseline deems critical ($\pi_{ref}(\bm{a}) > 0$ while $\pi_{\theta}(\bm{a}) \approx 0$). In the context of multi-vehicle collaboration, this perfectly models the risk of "fatal omissions"---where the neural network completely ignores a vital, highly collaborative, or safety-critical action required to resolve a complex bottleneck. Thus, $\mathcal{D}_{opt}$ acts as a strict detector for missing critical optimal decisions.

\subsubsection{The Bounded Optimality Score}
While $\mathcal{D}_{opt}$ is theoretically rigorous, its domain is $[0, \infty)$. An unbounded metric poses practical challenges for cross-scenario aggregation, training dynamic visualization, and hyperparameter tuning. To construct an intuitive diagnostic tool, we map this divergence into a bounded scalar.

\begin{definition}[Bounded Optimality Score]
The policy optimality score $\mathcal{M}_{opt}$ is formulated by applying a negative exponential mapping to the policy optimality distance:
\begin{equation}
	\begin{aligned}
		\mathcal{M}_{\text{opt}}(\pi_{\theta}|s) &\triangleq \exp\left(-\mathcal{D}_{\text{opt}}(\pi_{\theta}|s)\right) \\
		&= \exp\left(-D_{\text{KL}}\left(\pi_{ref}(\cdot|s) \parallel \pi_{\theta}(\cdot|s)\right)\right).
	\end{aligned}
	\label{eq:optimality_score}
\end{equation}
\end{definition}

\textbf{Remark (Interpretability for Diagnostics):} The exponential mapping elegantly projects the divergence into a normalized, strictly bounded range $\mathcal{M}_{opt} \in (0, 1]$. A score of $\mathcal{M}_{opt} = 1$ indicates that the MARL policy $\pi_{\theta}$ perfectly replicates the optimal synergy and action preferences of the MCTS baseline. Conversely, a score approaching $0$ signifies a catastrophic deviation from rational collaboration. This bounded nature enables researchers to robustly track the intrinsic quality of a policy as it evolves across thousands of training episodes, unaffected by the noise of extrinsic reward signals.

\subsection{Semantic Decoupling of Policy Dimensions}
\label{subsec:semantic_decoupling}

While the global optimality score $\mathcal{M}_{opt}$ provides a macroscopic evaluation of the joint policy, multi-vehicle collaboration inherently consists of two orthogonal physical tasks: spatial routing and velocity regulation. To achieve fine-grained diagnostics and pinpoint specific algorithmic deficiencies, we semantically decouple the holistic metric into distinct lateral and longitudinal dimensions.

\subsubsection{Action Space Decomposition and Marginalization}
In structured traffic environments, the individual action $a^{(i)}$ of any CAV $i$ can be factorized into a tuple $a^{(i)} = (a_{lat}^{(i)}, a_{lon}^{(i)})$. Here, $a_{lat}^{(i)} \in \mathcal{A}_{lat} = \{-1, 0, 1\}$ represents the lateral decisions (left lane-change, lane-keeping, right lane-change), and $a_{lon}^{(i)} \in \mathcal{A}_{lon} = \{-1, 0, 1\}$ denotes the longitudinal maneuvers (decelerate, maintain, accelerate)[cite: 1, 2].

To evaluate a specific dimension independently, we extract the marginal probability distributions by integrating (summing) out the orthogonal dimension. For the theoretical baseline $\pi_{\text{ref}}$, the marginal policy for agent $i$ along dimension $d \in \{\text{lat}, \text{lon}\}$ is derived as:
\begin{equation}
	\pi_{\text{ref}, d}^{(i)}(a_d^{(i)} | s) = \sum_{a_{d'} \in \mathcal{A}_{d'}} \sum_{\bm{a}^{-i} \in \mathcal{A}^{-i}} \pi_{\text{ref}}(a_d^{(i)}, a_{d'}, \bm{a}^{-i} | s),
	\label{eq:marginal_ref}
\end{equation}
where $d \in \{\text{lat}, \text{lon}\}, $ $d' \in \{\text{lat}, \text{lon}\} \setminus \{d\}$ denotes the complementary action dimension, and $\bm{a}^{-i}$ represents the joint action of all other agents except $i$. The marginal distributions for the evaluated MARL policy, $\pi_{\theta, d}^{(i)}$, are obtained analogously.

\subsubsection{Decoupled Optimality Metrics}
Using the extracted marginal distributions, we independently compute the KL divergence for each dimension and apply the bounded exponential mapping. 

\begin{definition}[Decoupled Optimality Scores]
The lateral policy optimality score $\mathcal{M}_{opt, lat}$ and the longitudinal policy optimality score $\mathcal{M}_{opt, lon}$ for the multi-vehicle system are defined as the mean decoupled scores across all agents:
\begin{equation}
		\mathcal{M}_{\text{opt}, d} = \exp\left( - \frac{1}{N} \sum_{i=1}^{N} D_{\text{KL}}\left(\pi_{\text{ref}, d}^{(i)}(\cdot|s) \parallel \pi_{\theta, d}^{(i)}(\cdot|s)\right) \right),
	\label{eq:decoupled_metrics}
\end{equation}
where $d \in \{\text{lat}, \text{lon}\}$.
\end{definition}

\textbf{Remark (Diagnostic Granularity for "Algorithmic Blindness"):} The decoupled metrics $\mathcal{M}_{opt, lat}$ and $\mathcal{M}_{opt, lon}$ equip the diagnostic framework with a powerful ``microscope.'' While a MARL algorithm might exhibit a seemingly acceptable global $\mathcal{M}_{opt}$, the decoupled scores can expose severe ``partial blindness'' in specific physical tasks. For instance, it can mathematically diagnose whether a structurally-driven exploration mechanism overly biases the network towards complex lateral lane-changes at the expense of fundamental longitudinal velocity coordination. This granular visibility is indispensable for evaluating specialized network architectures and targeted exploration rewards.
\section{Experimental Setup for Policy Diagnostics}
\label{sec:experimental_setup}

To systematically validate the proposed information-theoretic diagnostic framework, we construct a comprehensive evaluation matrix that stress-tests the multi-agent policies across varying degrees of environmental stochasticity and coordination complexity.

\subsection{Continuous Traffic Flow Simulation}
\label{subsec:traffic_simulation}

The diagnostic evaluation is conducted within a high-fidelity continuous traffic flow environment, modeled after a 250-meter segment of a four-lane urban arterial road with a design speed of 60 km/h. Unlike simplified grid-worlds or closed-loop tracks, the continuous injection and departure of vehicles mandate that the MARL policies robustly handle a time-varying number of agents and non-Euclidean interactive topologies.

To rigorously diagnose policy degradation boundaries, the evaluation matrix incorporates orthogonal combinations of two critical variables:

\textbf{Traffic Flow Rates:} We evaluate the algorithms under three representative flow rates: 400, 600, and 700 pcu/h/ln. These correspond to the Level of Service (LOS) C (stable flow), LOS D (high-density stable flow), and the LOS D/E boundary (critical near-congestion flow), respectively. This variation allows us to diagnose how algorithms compromise safety for efficiency under extreme congestion.

\textbf{CAV Penetration Rates:} For each flow rate, the policies are tested under 25\%, 50\%, 75\%, and 100\% CAV penetration rates. Lower penetration rates introduce severe partial observability and uncooperative uncertainties from background Human-Driven Vehicles (HDVs), thereby stress-testing the robustness of the MARL algorithms' spatial reasoning.

\subsection{The Oracle Baseline and Evaluated MARL Configurations}
\label{subsec:evaluated_configurations}

\subsubsection{The MCTS Oracle for Baseline Generation}
The theoretical ground-truth baseline policy, $\pi_{ref}(\cdot|s)$, is generated dynamically for a vast dataset of sampled traffic states using a highly optimized Parallel Evaluation MCTS (PE-MCTS) solver. Equipped with a parallel update pruning mechanism and experiential action preferences, the PE-MCTS acts as an oracle ($n_{\text{rollout}} = 1200$, $c_{\text{puct}} = 21$), generating a near-optimal value landscape $Q_{\text{MCTS}}^*(s, \bm{a})$ that guarantees strict asymptotic convergence.

\subsubsection{Evaluated MARL Algorithms}
To demonstrate the diagnostic versatility of the proposed metrics $\mathcal{M}_{opt}$, $\mathcal{M}_{opt, lat}$, and $\mathcal{M}_{opt, lon}$, we apply the framework to evaluate a wide spectrum of MARL architectures and exploration mechanisms. The evaluations are divided into two distinct diagnostic applications: \textbf{Spatial Representation Diagnostics:} To evaluate the capacity of different neural architectures in capturing dynamic multi-vehicle topologies, we compare the QMIX baseline integrated with four distinct agent networks: a Multi-Layer Perceptron (MLP), a Convolutional Neural Network (CNN), a Graph Attention Network (GAT), and our developed Representation Network (RepNet), which utilizes a Vision Transformer (ViT) backbone equipped with physical positional encoding. \textbf{Temporal Exploration Diagnostics:} To diagnose how different intrinsic reward mechanisms bias the learned policies, we evaluate state-of-the-art exploration algorithms, including CDS (diversity-driven)\cite{li2021cds}, SI2E (structural information-driven)\cite{zeng2024si2e}, MASER (subgoal-driven)\cite{jeon2022maser}, SPIE (retrospective information-driven)\cite{yu2023spie}, and our Topology-Enhanced (TPE)\cite{han2025tpe} algorithm, which utilizes game-theoretic topology and mutual information for guided collaboration.

\textbf{Remark (Standardization of Assessment):} All evaluated MARL algorithms adhere to the CTDE paradigm and are trained until convergence under identical environmental configurations. By holding the extrinsic reward structure constant across all experiments, any observed divergence in $\mathcal{M}_{opt}$ strictly isolates and quantifies the intrinsic architectural or exploratory deficiencies of the evaluated algorithms.

\section{Spatial Diagnostics: Uncovering Policy Biases}
\label{sec:spatial_diagnostics}

Traditional extrinsic metrics (e.g., success rate, collision rate) merely indicate macro-level task failures, offering zero transparency into the underlying causal mechanisms. By projecting the joint policy into the decoupled optimality space $(\mathcal{M}_{opt, lat}, \mathcal{M}_{opt, lon})$, our diagnostic framework acts as a semantic filter, mathematically exposing algorithmic blind spots in spatial reasoning.

\begin{figure}[htbp]
    \centering
    \begin{subfigure}{0.32\linewidth}
        \includegraphics[width=\linewidth]{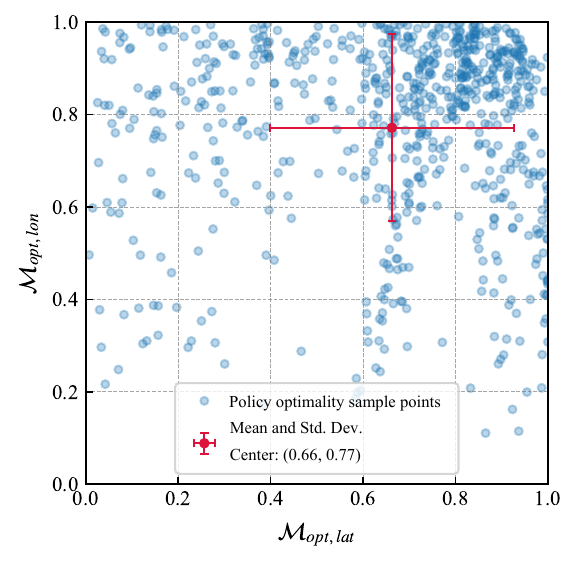}
        \caption{MLP}
        \label{sfig:diag_mlp}
    \end{subfigure}
    \hfill
    \begin{subfigure}{0.32\linewidth}
        \includegraphics[width=\linewidth]{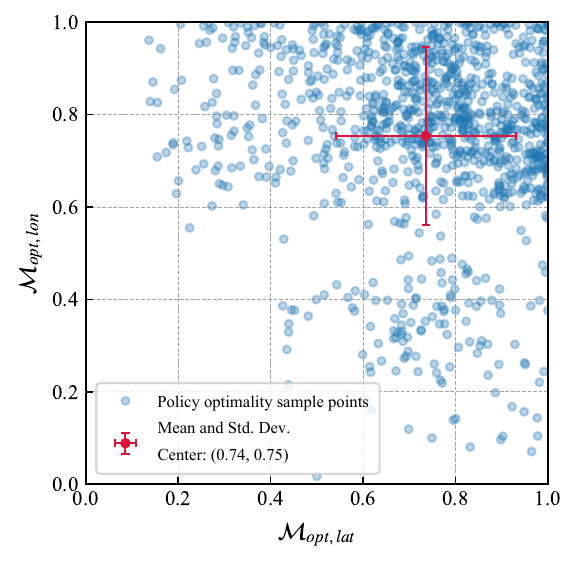}
        \caption{GAT}
        \label{sfig:diag_gat}
    \end{subfigure}
    \hfill
    \begin{subfigure}{0.32\linewidth}
        \includegraphics[width=\linewidth]{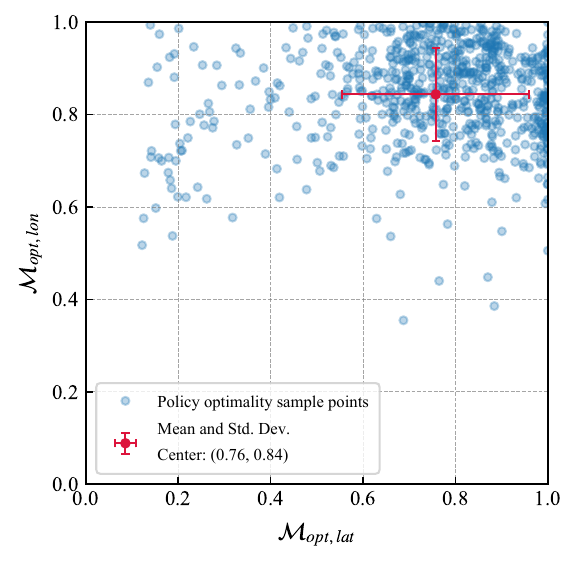}
        \caption{RepNet}
        \label{sfig:diag_repnet}
    \end{subfigure}
    \caption{Decoupled optimality distribution $(\mathcal{M}_{opt, lat}, \mathcal{M}_{opt, lon})$ for different multi-agent representation architectures. Red dots indicate the distribution mean center.}
    \label{fig:spatial_diag_rep}
\end{figure}

\subsection{Diagnosing Representation Network Architectures}
\label{subsec:diagnosing_representation}

The capability of an agent network to encode non-Euclidean vehicle interactions inherently dictates its policy ceiling. As illustrated in Fig. \ref{fig:spatial_diag_rep}, we diagnose three distinct architectures integrated within the QMIX framework: Multi-Layer Perceptron (MLP)\cite{rashid2018qmix}, Graph Attention Network (GAT), and ViT-based Representation Network (RepNet)\cite{han2024spformer}.

\textbf{The Failure of MLP:} As shown in Fig. \ref{fig:spatial_diag_rep}\subref{sfig:diag_mlp}, the MLP architecture exhibits severe degradation in both dimensions, with its optimality mean center collapsing to $(0.66, 0.77)$. This mathematically proves that simple feature concatenation is insufficient for resolving dynamic spatial interactions.

\textbf{Topological Limits of Graph Networks:} The GAT architecture demonstrates marginal improvements (Fig. \ref{fig:spatial_diag_rep}\subref{sfig:diag_gat}), achieving a mean center of $(0.74, 0.75)$. While explicitly modeling multi-vehicle interactions as graph edges aids in cross-lane relationship extraction (lateral planning), its longitudinal optimality remains highly constrained.

\textbf{Attention-Based Efficacy:} The RepNet architecture (Fig. \ref{fig:spatial_diag_rep}\subref{sfig:diag_repnet}) yields a significantly superior and balanced optimality distribution with a mean center of $(0.76, 0.84)$. This confirms that a transformer-based attention mechanism effectively captures both spatial and temporal dependencies required for robust cooperative planning.

\begin{figure}[htbp]
	\centering
	\begin{subfigure}{0.31\linewidth}
		\centering
		\includegraphics[width=\linewidth]{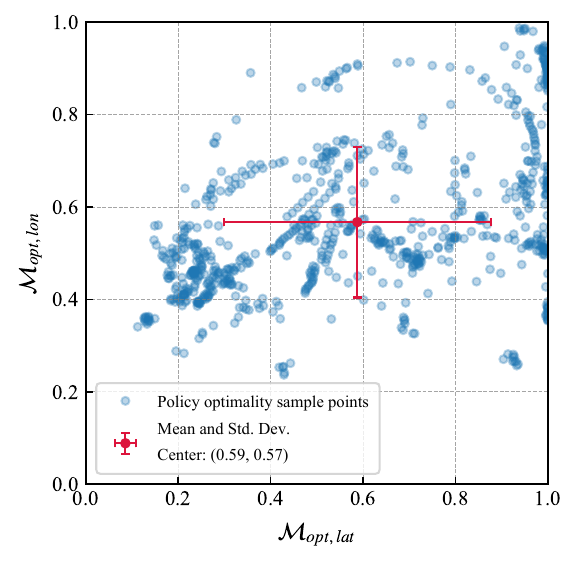}
		\caption{CDS}
		\label{sfig:diag_cds}
	\end{subfigure}
	\hfill
	\begin{subfigure}{0.31\linewidth}
		\centering
		\includegraphics[width=\linewidth]{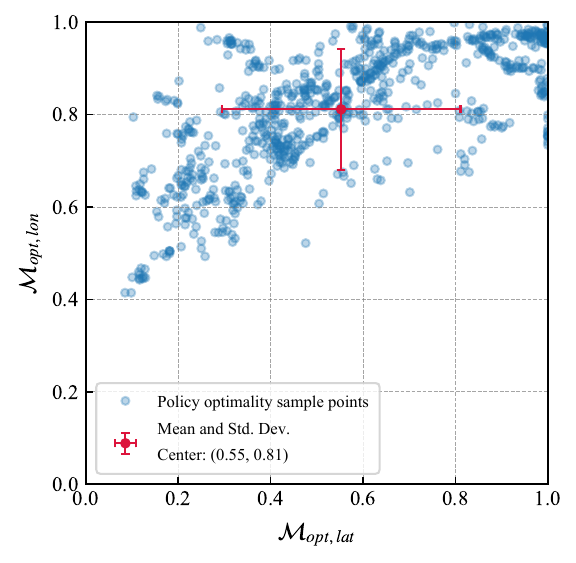}
		\caption{MASER}
		\label{sfig:diag_maser}
	\end{subfigure}
	\hfill
	\begin{subfigure}{0.31\linewidth}
		\centering
		\includegraphics[width=\linewidth]{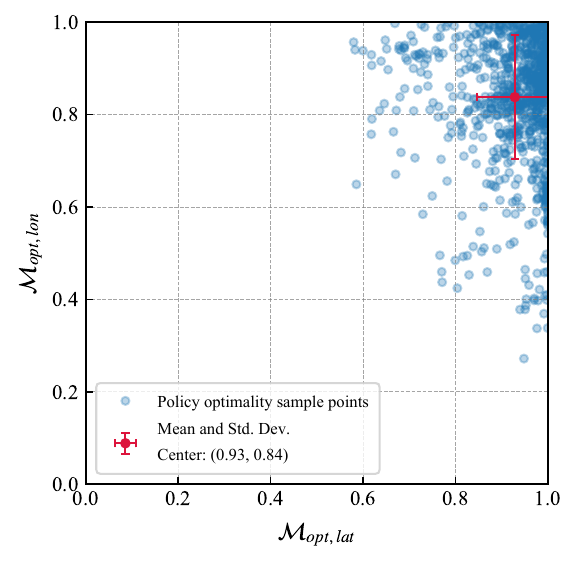}
		\caption{SI2E}
		\label{sfig:diag_si2e}
	\end{subfigure}
	
	\vspace{1.5ex} 
	
	\begin{subfigure}{0.31\linewidth}
		\centering
		\includegraphics[width=\linewidth]{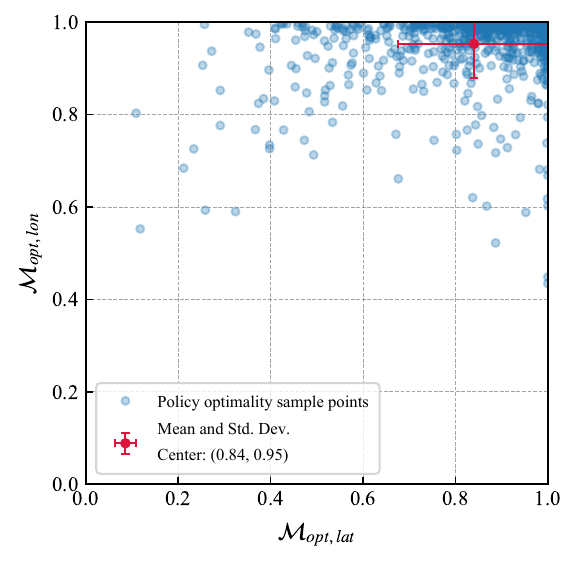} 
		\caption{SPIE}
		\label{sfig:diag_spie}
	\end{subfigure}
	\hspace{0.05\linewidth} 
	\begin{subfigure}{0.31\linewidth}
		\centering
		\includegraphics[width=\linewidth]{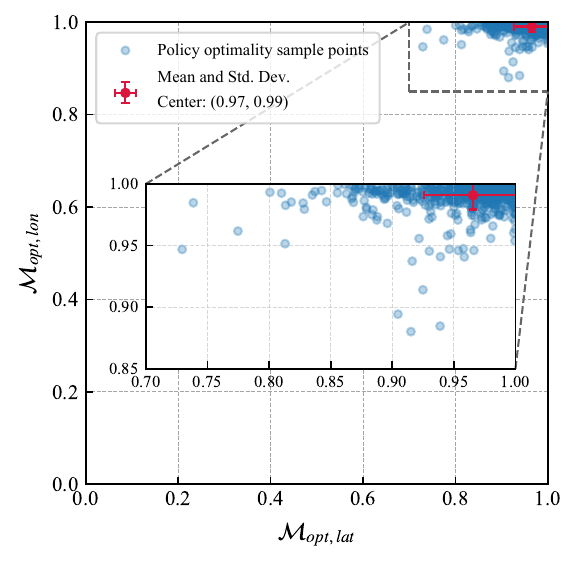}
		\caption{TPE}
		\label{sfig:diag_tpe}
	\end{subfigure}
	
	\caption{Decoupled optimality diagnosis for SOTA exploration mechanisms versus the proposed Topology-Enhanced (TPE) algorithm. The decoupled metric vividly uncovers the lateral or longitudinal biases inherent in unconstrained exploration paradigms.}
	\label{fig:spatial_diag_exp}
\end{figure}

\subsection{Identifying Biases in SOTA Exploration Algorithms}
\label{subsec:diagnosing_exploration}

Beyond network architectures, the design of intrinsic exploration rewards intrinsically biases the learned policy distribution. We apply the decoupled metric to diagnose four state-of-the-art (SOTA) exploration algorithms against our TPE baseline, as visualized in Fig. \ref{fig:spatial_diag_exp}. The diagnostics reveal profound directional biases:

\textbf{Lateral Bias in Structural Information (SI2E):} As shown in Fig. \ref{sfig:diag_si2e}, SI2E yields a highly asymmetric optimality distribution. It achieves a high lateral score ($\mathcal{M}_{opt, lat} \approx 0.93$), yet its longitudinal performance remains dispersed ($\mathcal{M}_{opt, lon} \approx 0.84$). This diagnoses a clear algorithmic bias: structurally-driven exploration heavily incentivizes discrete lateral lane-changes but fails to adequately regularize continuous longitudinal velocity coordination.

\textbf{Longitudinal Bias in Retrospective Exploration (SPIE):} Conversely, SPIE exhibits the exact inverse pathology (Fig. \ref{fig:spatial_diag_exp}\subref{sfig:diag_spie}). It achieves near-optimal longitudinal execution ($\mathcal{M}_{opt, lon} \approx 0.95$) but struggles laterally ($\mathcal{M}_{opt, lat} \approx 0.84$). This mathematically isolates the limitation of retrospective information tracking: it excels in optimizing temporally coherent car-following behavior but lacks the spatial horizon necessary for complex lateral negotiation.

\textbf{Lateral Failure in Subgoal-Driven Methods (MASER):} MASER exhibits a highly dispersed distribution with a mean center of $(0.55, 0.81)$ (Fig. \ref{fig:spatial_diag_exp}\subref{sfig:diag_maser}). While subgoal generation provides some longitudinal guidance, the low lateral score explicitly correlates with its empirically observed low success rates, proving its inadequacy in multi-lane bottleneck resolution.

\textbf{The Pitfall of Unconstrained Diversity (CDS):} The CDS algorithm collapses entirely, yielding a mean center of $(0.59, 0.57)$ (Fig. \ref{fig:spatial_diag_exp}\subref{sfig:diag_cds}). The metric proves that unconstrained diversity-driven exploration disperses computational resources indiscriminately, preventing convergence to any optimal collaborative policy in either dimension.

\textbf{Remark (Diagnostic Conclusion):} In stark contrast to the SOTA baselines, the proposed TPE mechanism achieves a concentrated optimality mean of $(0.97, 0.99)$ (Fig. \ref{fig:spatial_diag_exp}\subref{sfig:diag_tpe}). The visual and numerical evidence provided by the decoupled metrics conclusively proves that only a balanced, topology-aware exploration strategy can eliminate directional biases and converge to the true theoretical baseline.
\section{Temporal Diagnostics: Evaluating Dynamic Adaptation}
\label{sec:temporal_diagnostics}

Macro-level reward convergence often masks temporal policy degradation. MARL agents can collapse into static, average behaviors that perform adequately in trivial states but fail during complex negotiations. We utilize time-series action heatmaps to diagnose this dynamic adaptation deficiency against the MCTS oracle.

\begin{figure}[htbp]
    \centering
    \begin{subfigure}{0.48\linewidth}
        \includegraphics[width=\linewidth]{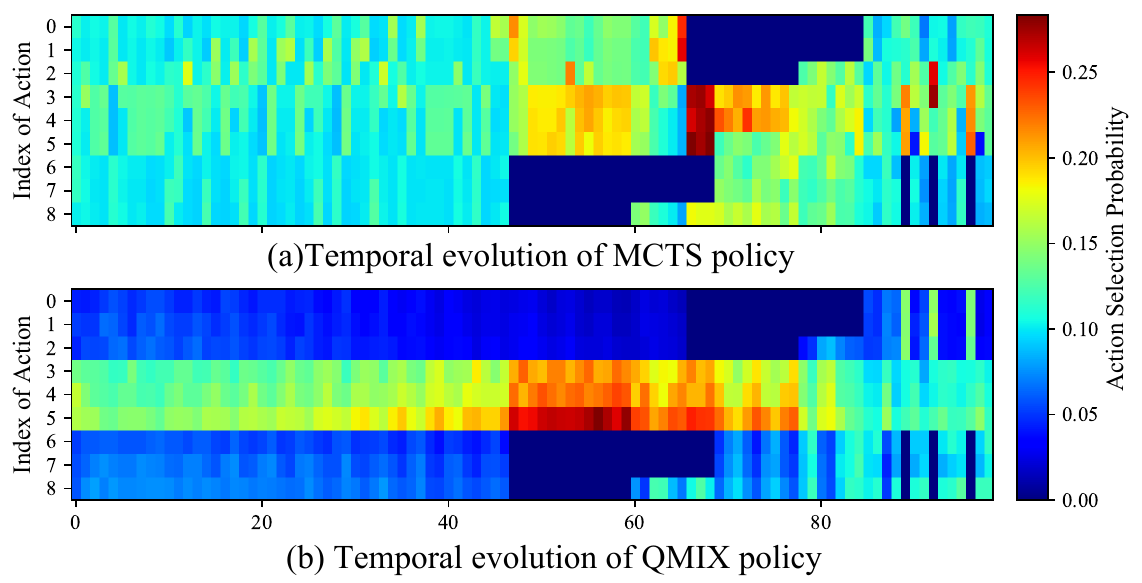}
        \caption{RepNet-QMIX}
        \label{sfig:temp_single_repnet}
    \end{subfigure}
    \hfill
    \begin{subfigure}{0.48\linewidth}
        \includegraphics[width=\linewidth]{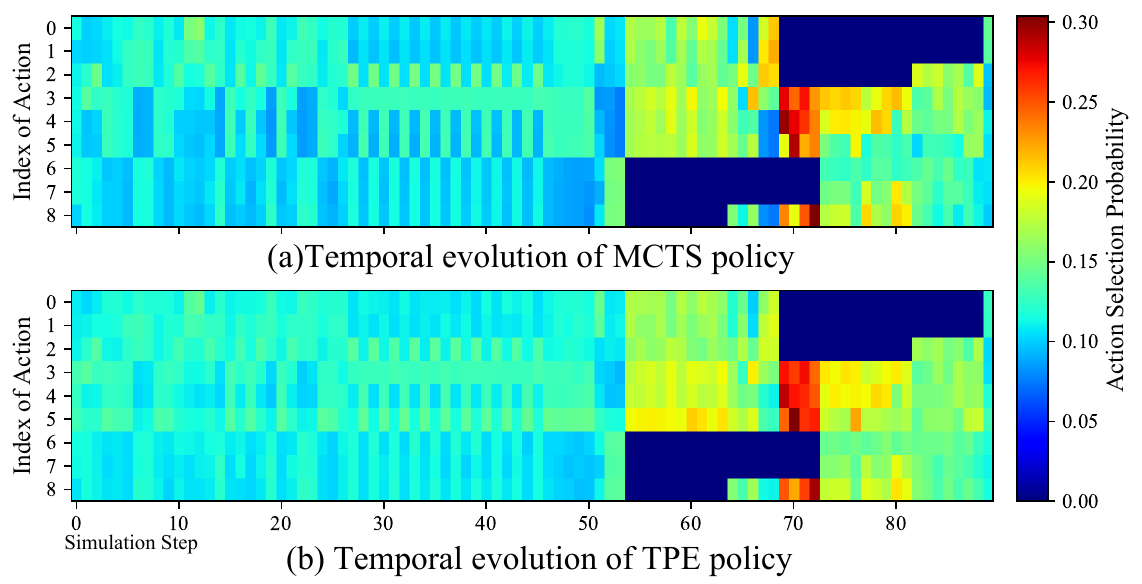}
        \caption{TPE}
        \label{sfig:temp_single_tpe}
    \end{subfigure}
    \caption{Temporal evolution of single-agent policies. The top rows represent the dynamic MCTS baseline, while the bottom rows show the evaluated MARL policy.}
    \label{fig:temp_single}
\end{figure}

\subsection{Identifying the ``Average Policy'' Trap}
\label{subsec:average_policy_trap}

Visualizing the single-agent policy evolution (Fig. \ref{fig:temp_single}) exposes the behavioral rigidity of standard representation networks.

\textbf{The RepNet Collapse:} As shown in Fig. \ref{fig:temp_single}\subref{sfig:temp_single_repnet}, the RepNet-QMIX policy stagnates into an average policy trap. Its probability mass anchors continuously on Action 4 (maintain speed) and Action 5 (accelerate). It completely fails to track the dynamic state-dependent switching of the MCTS baseline, yielding a mere $33.94\%$ single-action match rate.

\textbf{TPE Dynamic Tracking:} Conversely, the TPE algorithm (Fig. \ref{fig:temp_single}\subref{sfig:temp_single_tpe}) dynamically shifts its probability hotspots in strict alignment with the MCTS oracle. The topology-enhanced exploration effectively prevents behavioral stagnation, elevating the single-action match rate significantly to $69.49\%$.

\begin{figure}[htbp]
    \centering
    \begin{subfigure}{0.48\linewidth}
        \includegraphics[width=\linewidth]{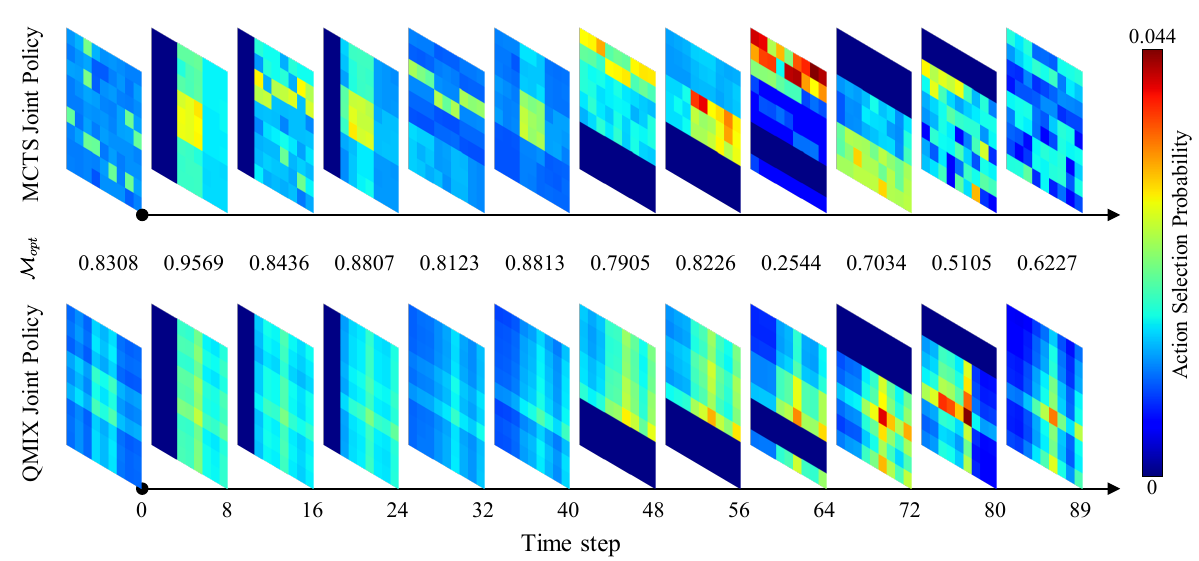}
        \caption{RepNet-QMIX (Joint Policy)}
        \label{sfig:temp_joint_repnet}
    \end{subfigure}
    \hfill
    \begin{subfigure}{0.48\linewidth}
        \includegraphics[width=\linewidth]{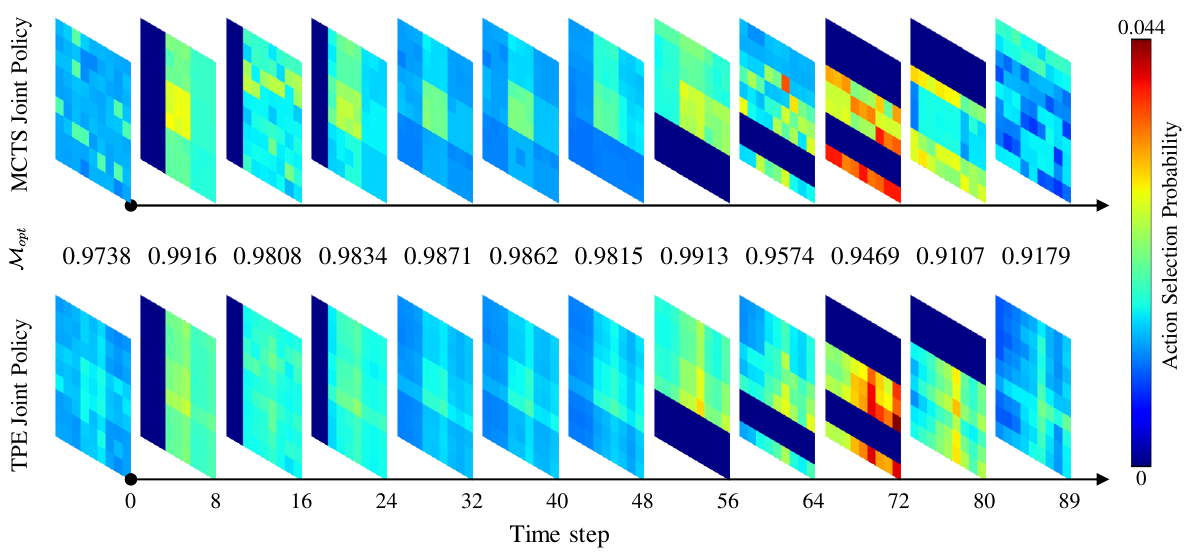}
        \caption{TPE (Joint Policy)}
        \label{sfig:temp_joint_tpe}
    \end{subfigure}
    \caption{Temporal evolution of joint policies alongside the continuous tracking of the global optimality score $\mathcal{M}_{opt}$.}
    \label{fig:temp_joint}
\end{figure}

\subsection{Quantifying Joint Synergy Fluctuation}
\label{subsec:joint_synergy}

The temporal tracking of the joint policy (Fig. \ref{fig:temp_joint}) provides a rigorous assessment of multi-agent synergy.

\textbf{Volatility in Standard MARL:} Fig. \ref{fig:temp_joint}\subref{sfig:temp_joint_repnet} reveals severe structural discrepancies in the RepNet-QMIX joint policy. Its optimality score $\mathcal{M}_{opt}$ fluctuates violently between $0.2544$ and $0.9569$, proving that its occasional high returns are coincidental rather than systematically synergistic. Consequently, its optimal joint action match rate is an abysmal $6.74\%$.

\textbf{Synergistic Stability in TPE:} The TPE joint policy (Fig. \ref{fig:temp_joint}\subref{sfig:temp_joint_tpe}) exhibits high structural fidelity to the MCTS baseline. Its $\mathcal{M}_{opt}$ strictly stabilizes above $0.91$ throughout the entire episodic horizon. The optimal joint action match rate improves significantly to $15.73\%$, confirming that the intrinsic mutual information reward successfully binds individual decisions into coherent, long-horizon collaborative strategies.

\section{Optimality-Guided Hyperparameter Optimization}
\label{sec:hyperparameter_optimization}

Traditional hyperparameter tuning relies on extrinsic reward curves, which suffer from high variance and sparse signals, often masking the structural deterioration of policies. By mapping the policy quality into a bounded 2D metric space, our framework transforms hyperparameter tuning into a visual, vector-based trajectory tracking process.

\begin{figure}[htbp]
    \centering
    \begin{subfigure}{0.48\linewidth}
        \includegraphics[width=\linewidth]{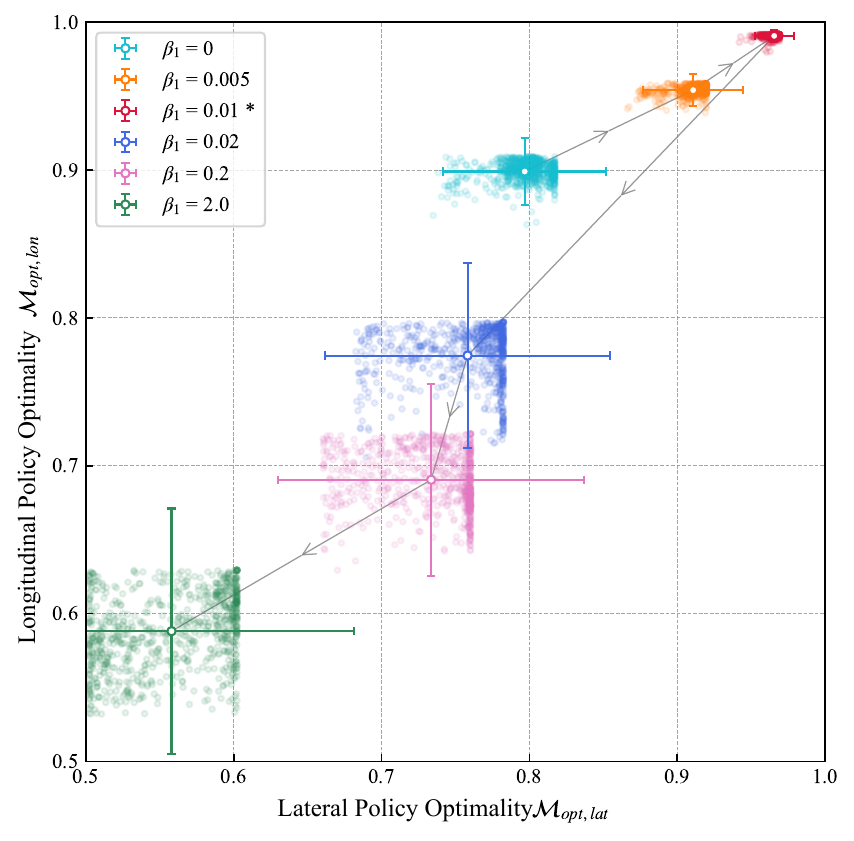}
        \caption{Trajectory of $\beta_1$}
        \label{sfig:param_beta1}
    \end{subfigure}
    \hfill
    \begin{subfigure}{0.48\linewidth}
        \includegraphics[width=\linewidth]{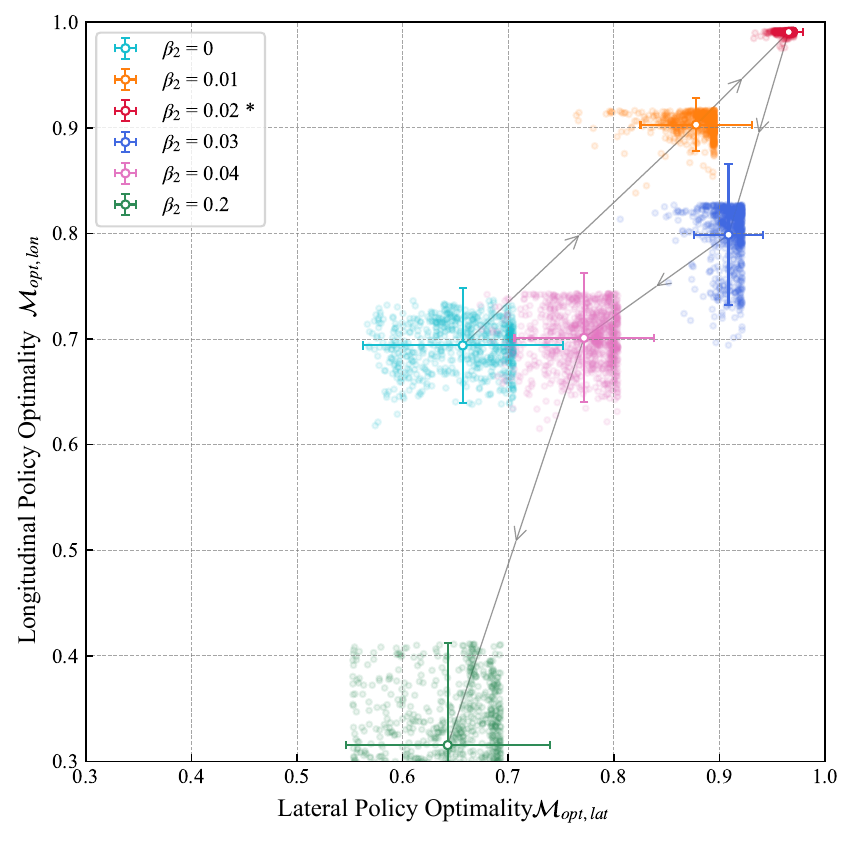}
        \caption{Trajectory of $\beta_2$}
        \label{sfig:param_beta2}
    \end{subfigure}
    \caption{Trajectory tracking of the optimality mean centers under varying intrinsic reward weights. The optimal configurations mathematically converge toward the $(1, 1)$ coordinate.}
    \label{fig:param_opti}
\end{figure}

\subsection{Tracking the Trajectory of Metric Distribution Centers}
\label{subsec:tracking_trajectory}

We apply the decoupled metric to diagnose the sensitivity of the TPE algorithm to its two intrinsic reward weights: topological novelty ($\beta_1$) and mutual information-based collaboration ($\beta_2$).

\textbf{Balancing Exploration Breadth ($\beta_1$):} As shown in Fig. \ref{fig:param_opti}\subref{sfig:param_beta1}, increasing $\beta_1$ from $0$ to $0.01$ pulls the optimality mean center drastically from $(0.8, 0.9)$ to $(0.97, 0.99)$, achieving peak policy quality. However, excessive weight ($\beta_1 \to 2.0$) induces a severe regression toward the lower-left quadrant. The metric mathematically diagnoses the over-exploration collapse: the algorithm abandons fundamental driving tasks, becoming entirely distracted by the pursuit of novel but meaningless topological states.

\textbf{Balancing Collaborative Exploitation ($\beta_2$):} The trajectory of $\beta_2$ (Fig. \ref{fig:param_opti}\subref{sfig:param_beta2}) provides even deeper diagnostic insights. Increasing $\beta_2$ from $0$ to $0.02$ successfully shifts the center from $(0.65, 0.70)$ to $(0.93, 0.84)$. Yet, an excessive weight ($\beta_2 \ge 0.2$) triggers a sharp downward drift in longitudinal optimality. This visually captures a collusion trap: agents learn to minimize speed and remain static to artificially stabilize their mutual topologies (maximizing predictability and mutual information), completely sacrificing the extrinsic throughput objective.

\section{Conclusion}
\label{sec:conclusion}

This paper introduces an information-theoretic diagnostic framework that transitions MARL evaluation in autonomous driving from extrinsic statistical indicators to intrinsic policy quality. By establishing a rigorous MCTS theoretical baseline and decoupling the KL divergence into lateral and longitudinal dimensions, the proposed metrics ($\mathcal{M}_{opt, lat}$, $\mathcal{M}_{opt, lon}$) function as a semantic microscope. The extensive spatial and temporal diagnostics conclusively prove the framework's capability to expose algorithmic blind spots, identify average-policy traps, and guide optimal hyperparameter configurations. This framework offers a robust, model-agnostic standard for benchmarking future multi-agent robotic systems.

\bibliographystyle{IEEEtran.bst}
\bibliography{ref.bib}

@article{han2025tpe,
      title={Topology Enhanced MARL for Multi-Agent Cooperative Decision-Making of CAVs}, 
      author={Ye Han and Lijun Zhang and Dejian Meng and Zhuang Zhang},
      year={2025},
      journal={arXiv preprint arXiv:2507.12110}, 
}

@INPROCEEDINGS{han2024spformer,
  author={Han, Ye and Zhang, Lijun and Meng, Dejian and Hu, Xingyu and Lu, Yixia},
  booktitle={Proc. IEEE Int. Conf. Intell. Transp. Syst.}, 
  title={SPformer: A Transformer Based DRL Decision Making Method for Connected Automated Vehicles}, 
  year={2024},
  volume={},
  number={},
  pages={1223-1230},
  }

@ARTICLE{han2025mcts,
  author={Han, Ye and Zhang, Lijun and Meng, Dejian and Zhang, Zhuang and Hu, Xingyu and Weng, Songyu},
  journal = {IEEE Trans. Intell. Transp. Syst.}, 
  title={A Value-Based Parallel Update MCTS Method for Multi-Agent Cooperative Decision-Making of Connected and Automated Vehicles}, 
  year={2026},
  volume={27},
  number={1},
  pages={1400-1415},
}

@inproceedings{rashid2018qmix,
  title     = {QMIX: Monotonic Value Function Factorisation for Deep Multi-Agent Reinforcement Learning},
  author    = {Rashid, Tabish and Samvelyan, Mikayel and Schroeder, Christian and Farquhar, Gregory and Foerster, Jakob and Whiteson, Shimon},
  booktitle = {Proc. 35th Int. Conf. Mach. Learn.},
  volume    = {80},
  pages     = {4295--4304},
  year      = {2018}
}

@inproceedings{li2021cds,
  author    = {Li, Chenghao and Wang, Tonghan and Wu, Chengjie and Zhao, Qianchuan and Yang, Jun and Zhang, Chongjie},
  booktitle = {Adv. Neural Inf. Process. Syst.},
  pages     = {3991--4002},
  title     = {Celebrating Diversity in Shared Multi-Agent Reinforcement Learning},
  volume    = {34},
  year      = {2021}
}

@inproceedings{yu2023spie,
  author    = {Yu, Changmin and Burgess, Neil and Sahani, Maneesh and Gershman, Samuel J.},
  title     = {Successor-Predecessor Intrinsic Exploration},
  booktitle = {Adv. Neural Inf. Process. Syst.},
  year      = {2023}
}

@inproceedings{zeng2024si2e,
  author    = {Zeng, Xianghua and Peng, Hao and Li, Angsheng},
  title     = {Effective Exploration Based on the Structural Information Principles},
  booktitle = {Adv. Neural Inf. Process. Syst.},
  year      = {2024}
}

@inproceedings{jeon2022maser,
  title     = {{MASER}: Multi-Agent Reinforcement Learning with Subgoals Generated from Experience Replay Buffer},
  author    = {Jeon, Jeewon and Kim, Woojun and Jung, Whiyoung and Sung, Youngchul},
  booktitle = {Proc. 39th Int. Conf. Mach. Learn.},
  pages     = {10041--10052},
  year      = {2022},
  volume    = {162},
  month     = {17--23 Jul}
}

@article{chen2021graph,
  title={Graph neural network and reinforcement learning for multi-agent cooperative control of connected autonomous vehicles},
  author={Chen, Sikai and Dong, Jiqian and Ha, Paul Young Joun and Li, Yujie and Labi, Samuel},
  journal={Comput. Aided Civ. Inf. Eng.},
  volume={36},
  number={9},
  pages={1183--1197},
  year={2021}
}

@article{hua2024multi,
  title={Multi-Agent Reinforcement Learning for Connected and Automated Vehicles Control: Recent Advancements and Future Prospects},
  author={Hua, Min and Qi, Xinda and Chen, Dong and Jiang, Kun and Liu, Zemin Eitan and Sun, Hongyu and Zhou, Quan and Xu, Hongming},
  journal={IEEE Trans. Intell. Transp. Syst.},
  year={2024}
}

@article{wang2024multi,
  title={Multi-Agent DRL-Controlled Connected and Automated Vehicles in Mixed Traffic With Time Delays},
  author={Wang, Zhuwei and Xue, Yi and Liu, Lihan and Zhang, Haijun and Qu, Chunhui and Fang, Chao},
  journal={IEEE Trans. Intell. Transp. Syst.},
  year={2024}
}

@article{guillen2022multi,
  title={Multi-Agent Deep Reinforcement Learning to Manage Connected Autonomous Vehicles at Tomorrow's Intersections},
  author={Guillen-Perez, Antonio and Cano, Maria-Dolores},
  journal={IEEE Trans. Intell. Transp. Syst.},
  year={2022}
}

@article{xu2024multi,
  title={A Multi-Agent Reinforcement Learning Based Control Method for CAVs in a Mixed Platoon},
  author={Xu, Yaqi and Shi, Yan and Tong, Xiaolu and Chen, Shanzhi and Ge, Yuming},
  journal={IEEE Trans. Intell. Veh.},
  year={2024}
}

@article{guo2024heuristic,
  title={Heuristic-Based Multi-Agent Deep Reinforcement Learning Approach for Coordinating Connected and Automated Vehicles at Non-Signalized Intersection},
  author={Guo, Zihan and Wu, Yan and Wang, Lifang and Zhang, Junzhi},
  journal={IEEE Trans. Veh. Technol.},
  year={2024}
}

@article{chen2023deep,
  title={Deep Multi-Agent Reinforcement Learning for Highway On-Ramp Merging in Mixed Traffic},
  author={Chen, Dong and Hajidavalloo, Mohammad R and Li, Zhaojian and Chen, Kaian and Wang, Yongqiang and Jiang, Longsheng and Wang, Yue},
  journal={IEEE Trans. Intell. Transp. Syst.},
  year={2023}
}

@article{saleem2024multi,
  title={Multi-Agent Reinforcement Learning for Real-Time Adaptive Lane Control in Mixed CAV-HDV Freeways Traffic},
  author={Saleem, Muhammad Asim and Zhou, Shijie and Quraishi, Aadam and Shabaz, Mohammad and Javed, Iram and Basheer, Shakila and Alenezi, Abdullah Feraih and Aldawsari, Hamad},
  journal={IEEE Trans. Consum. Electron.},
  year={2024}
}

@article{han2022multi,
  title={A Multi-Agent Reinforcement Learning Approach for Safe and Efficient Behavior Planning of Connected Autonomous Vehicles},
  author={Han, Songyang and Zhou, Shanglin and Wang, Jiangwei and Pepin, Lynn and Ding, Caiwen and Fu, Jie and Miao, Fei},
  journal={IEEE Trans. Intell. Transp. Syst.},
  year={2022}
}

@article{wang2023multi,
  title={A multi-agent reinforcement learning-based longitudinal and lateral control of CAVs to improve traffic efficiency in a mandatory lane change scenario},
  author={Wang, Shupei and Wang, Ziyang and Jiang, Rui and Zhu, Feng and Yan, Ruidong and Shang, Ying},
  journal={Transp. Res. Part C Emerg. Technol.},
  year={2023}
}

@article{zhang2023spatial,
  title={Spatial-Temporal-Aware Safe Multi-Agent Reinforcement Learning of Connected Autonomous Vehicles in Challenging Scenarios},
  author={Zhang, Zhili and Han, Songyang and Wang, Jiangwei and Miao, Fei},
  journal={IEEE Trans. Intell. Transp. Syst.},
  year={2023}
}

@inproceedings{henderson2018deep,
  title={Deep Reinforcement Learning that Matters},
  author={Henderson, Peter and Islam, Riashat and Bachman, Philip and Pineau, Joelle and Precup, Doina and Meger, David},
  booktitle={Proc. AAAI Conf. Artif. Intell. (AAAI)},
  volume={32},
  number={1},
  year={2018}
}

@inproceedings{li2025hierarchical,
  title={Hierarchical Reward Function Design for Autonomous Driving Based on Reinforcement Learning},
  author={Li, Yansong},
  booktitle={Proc. Int. Symp. Machine Learning and Social Computing (MLSC)},
  year={2025}
}

@inproceedings{liu2025rrm,
  title={RRM: Robust Reward Model Training Mitigates Reward Hacking},
  author={Liu, Tianqi and Xiong, Wei and Ren, Jie and Chen, Lichang and Wu, Junru and Joshi, Rishabh and Gao, Yang and Shen, Jiaming and Qin, Zhen and Yu, Tianhe and Sohn, Daniel and Makarova, Anastasiia and Liu, Jeremiah and Liu, Yuan and Piot, Bilal and Ittycheriah, Abe and Kumar, Aviral and Saleh, Mohammad},
  booktitle={Proc. Int. Conf. Learn. Represent. (ICLR)},
  year={2025}
}

@article{abouelazm2024review,
  title={A Review of Reward Functions for Reinforcement Learning in the context of Autonomous Driving},
  author={Abouelazm, Ahmed and Michel, Jonas and Zollner, J Marius},
  journal={arXiv preprint arXiv:2404.18520},
  year={2024}
}

@article{jordan2020evaluating,
  title={Evaluating the Performance of Reinforcement Learning Algorithms},
  author={Jordan, Scott M and Chandak, Yash and Cohen, Daniel and Zhang, Mengxue and Thomas, Philip S},
  journal={J. Mach. Learn. Res. (JMLR)},
  volume={21},
  pages={1--34},
  year={2020}
}

@article{guo2025opencda,
  title={OpenCDA-MARL: A Unified Benchmarking Framework for Cooperative Autonomous Intersection Management With Multi-Agent Reinforcement Learning},
  author={Guo, Lihao and Liu, Louis and Tang, Jiahao and Liu, Bo and Cao, Siyang},
  journal={IEEE Trans. Intell. Transp. Syst.},
  year={2025}
}

@article{yan2024policy,
  title={Policy Evaluation and Seeking for Multiagent Reinforcement Learning via Best Response},
  author={Yan, Rui and Duan, Xiaoming and Shi, Zongying and Zhong, Yisheng and Marden, Jason R and Bullo, Francesco},
  journal={IEEE Trans. Autom. Control},
  year={2024}
}

@article{glanois2024survey,
  title={A survey on interpretable reinforcement learning},
  author={Glanois, Claire and Weng, Paul and Zimmer, Matthieu and Li, Dong and Yang, Tianpei and Hao, Jianye and Liu, Wulong},
  journal={Mach. Learn.},
  volume={113},
  number={8},
  pages={5229--5288},
  year={2024}
}

@inproceedings{zhou2020smarts,
  title={SMARTS: Scalable Multi-Agent Reinforcement Learning Training School for Autonomous Driving},
  author={Zhou, Ming and Luo, Jun and Villella, Julian and Yang, Yaodong and Rusu, David and Miao, Jiayu and Zhang, Weinan and Alban, Montgomery and Fadakar, Iman and Chen, Zheng and others},
  booktitle={Proc. Conf. Robot Learn. (CoRL)},
  pages={264--285},
  year={2020}
}

@article{silver2017mastering,
  title={Mastering Chess and Shogi by Self-Play with a General Reinforcement Learning Algorithm},
  author={Silver, David and Hubert, Thomas and Schrittwieser, Julian and Antonoglou, Ioannis and Lai, Matthew and Guez, Arthur and Lanctot, Marc and Sifre, Laurent and Kumaran, Dharshan and Graepel, Thore and Lillicrap, Timothy and Simonyan, Karen and Hassabis, Demis},
  journal={arXiv preprint arXiv:1712.01815},
  year={2017}
}

@article{helfenstein2024checkmating,
  title={Checkmating One, by Using Many: Combining Mixture of Experts With MCTS to Improve in Chess},
  author={Helfenstein, Felix and Czech, Johannes and Bl{\"u}ml, Jannis and Eisel, Max and Kersting, Kristian},
  journal={IEEE Trans. Games},
  year={2024}
}

@article{swiechowski2023monte,
  title={Monte Carlo Tree Search: a review of recent modifications and applications},
  author={{\'{S}}wiechowski, Maciej and Godlewski, Konrad and Sawicki, Bartosz and Ma{\'{n}}dziuk, Jacek},
  journal={Artif. Intell. Rev.},
  volume={56},
  number={3},
  pages={2497--2562},
  year={2023}
}

@article{vinyals2019grandmaster,
  title={Grandmaster level in StarCraft II using multi-agent reinforcement learning},
  author={Vinyals, Oriol and Babuschkin, Igor and Czarnecki, Wojciech M. and Mathieu, Micha{\"{e}}l and Dudzik, Andrew and Chung, Junyoung and Choi, David H. and Powell, Richard and Ewalds, Timo and Georgiev, Petko and Oh, Junhyuk and Horgan, Dan and Kroiss, Manuel and Danihelka, Ivo and Huang, Aja and Sifre, Laurent and Cai, Trevor and Agapiou, John P. and Jaderberg, Max and Vezhnevets, Alexander S. and Leblond, R{\'{e}}mi and Pohlen, Tobias and Dalibard, Valentin and Budden, David and Sulsky, Yury and Molloy, James and Paine, Tom L. and Gulcehre, Caglar and Wang, Ziyu and Pfaff, Tobias and Wu, Yuhuai and Ring, Roman and Yogatama, Dani and W{\"{u}}nsch, Dario and McKinney, Katrina and Smith, Oliver and Schaul, Tom and Lillicrap, Timothy and Kavukcuoglu, Koray and Hassabis, Demis and Apps, Chris and Silver, David},
  journal={Nature},
  volume={575},
  number={7782},
  pages={350--354},
  year={2019}
}

@article{chan2022greedification,
  title={Greedification Operators for Policy Optimization: Investigating Forward and Reverse KL Divergences},
  author={Chan, Alan and Silva, Hugo and Lim, Sungsu and Kozuno, Tadashi and Mahmood, A. Rupam and White, Martha},
  journal={J. Mach. Learn. Res.},
  volume={23},
  pages={1--61},
  year={2022}
}

@article{schulman2017proximal,
  title={Proximal Policy Optimization Algorithms},
  author={Schulman, John and Wolski, Filip and Dhariwal, Prafulla and Radford, Alec and Klimov, Oleg},
  journal={arXiv preprint arXiv:1707.06347},
  year={2017}
}

@inproceedings{dou2024measuring,
  title={Measuring Mutual Policy Divergence for Multi-Agent Sequential Exploration},
  author={Dou, Haowen and Dang, Lujuan and Luan, Zhirong and Chen, Badong},
  booktitle={Proc. Adv. Neural Inf. Process. Syst. (NeurIPS)},
  year={2024}
}

@inproceedings{schulman2015trust,
  title={Trust Region Policy Optimization},
  author={Schulman, John and Levine, Sergey and Moritz, Philipp and Jordan, Michael and Abbeel, Pieter},
  booktitle={Proc. Int. Conf. Mach. Learn. (ICML)},
  pages={1889--1897},
  year={2015}
}

\end{document}